\documentclass[10pt,twocolumn]{article}

\usepackage[letterpaper,top=0.9in,bottom=0.9in,left=0.75in,right=0.75in,
            columnsep=0.28in]{geometry}
\usepackage[T1]{fontenc}
\usepackage[utf8]{inputenc}
\usepackage{mathptmx}          
\usepackage{amsmath,amssymb}
\usepackage{graphicx}
\usepackage{enumitem}
\usepackage{microtype}
\usepackage{setspace}
\usepackage{stfloats}          
\usepackage[skip=5pt,font=small,labelfont=bf,justification=justified,
            singlelinecheck=false]{caption}
\usepackage[most]{tcolorbox}
\usepackage{titlesec}
\usepackage{changepage}
\usepackage[colorlinks=true,allcolors=blue!55!black,breaklinks=true]{hyperref}
\usepackage{url}

\usepackage[backend=biber,style=nature,sorting=none,intitle=true]{biblatex}
\DeclareFieldFormat[inbook,incollection,inproceedings]{title}{#1}
\DeclareFieldFormat[misc,online,unpublished]{title}{#1}
\renewbibmacro*{in:}{\printtext{\bibcpstring{in}\intitlepunct}}

\titleformat{\section}{\normalfont\large\bfseries}{}{0pt}{}
\titlespacing*{\section}{0pt}{10pt plus 2pt minus 2pt}{4pt}
\titleformat{\subsection}{\normalfont\normalsize\bfseries\itshape}{}{0pt}{}
\titlespacing*{\subsection}{0pt}{8pt plus 2pt minus 2pt}{3pt}

\definecolor{boxframe}{HTML}{2C4A72}
\definecolor{boxback}{HTML}{F4F7FB}
\definecolor{subframe}{HTML}{7C93B3}
\definecolor{takeback}{HTML}{EFF4FA}

\newtcolorbox{paperbox}[1]{%
  breakable, enhanced,
  colback=boxback, colframe=boxframe,
  boxrule=1pt, arc=3pt,
  left=7pt, right=7pt, top=7pt, bottom=7pt,
  fonttitle=\bfseries\normalsize, coltitle=white, colbacktitle=boxframe,
  title=#1, toptitle=3pt, bottomtitle=3pt,
  before skip=10pt, after skip=10pt,
  bottomrule at break=0pt, toprule at break=0pt,
}

\newtcolorbox{subbox}[1][]{%
  breakable, enhanced,
  colback=white, colframe=subframe,
  boxrule=0.6pt, arc=2pt,
  left=6pt, right=6pt, top=5pt, bottom=5pt,
  fonttitle=\bfseries\small, coltitle=boxframe, colbacktitle=white,
  title=#1, toptitle=2pt, bottomtitle=0pt,
  before skip=6pt, after skip=6pt,
  bottomrule at break=0pt, toprule at break=0pt,
}

\newtcolorbox{takeawaybox}{%
  breakable, enhanced,
  colback=takeback, colframe=boxframe,
  boxrule=0.6pt, arc=2pt,
  left=6pt, right=6pt, top=5pt, bottom=5pt,
  before skip=6pt, after skip=2pt,
  bottomrule at break=0pt, toprule at break=0pt,
}

\newcommand{\qitem}[2]{\par\smallskip\noindent\textbf{#1}\par\nobreak
  \vspace{1pt}\noindent #2\par}

\begin{document}

\twocolumn[
\begin{@twocolumnfalse}
\begin{center}
{\LARGE\bfseries What Can Analogy Tell Us About\\[2pt]
Artificial Consciousness?\par}
\vspace{10pt}
{\large Keith J. Holyoak\textsuperscript{1,$^\star$}, \ Martin M. Monti\textsuperscript{1,2}\par}
\vspace{6pt}
{\small
\textsuperscript{1}Department of Psychology, University of California, Los Angeles\\
\textsuperscript{2}Brain Injury Research Center, Department of Neurosurgery,
University of California, Los Angeles\par}
\vspace{4pt}
{\small $^\star$ Correspondence: Keith Holyoak, Department of Psychology, University of
California Los Angeles, Los Angeles CA 90095\\
\href{mailto:holyoak@ucla.edu}{kholyoak@g.ucla.edu},
\href{mailto:monti@ucla.edu}{monti@g.ucla.edu}\par}
\end{center}
\vspace{6pt}
\begin{adjustwidth}{0.06\textwidth}{0.06\textwidth}
\noindent\textbf{Abstract.} Who or what is conscious? Because subjective experience is directly
accessible only in the first person, judgments about consciousness in
other entities depend partly on analogy. Historically, such inferences
have focused on nonhuman animals, but advances in artificial
intelligence have raised the possibility of conscious AI. Here we
develop a causal framework for evaluating such evidential analogies. The
key distinction is between similarities in factors plausibly involved in
generating consciousness and similarities in downstream behavioural or
cognitive effects. Our framework weights source--target similarity by
causal relevance while allowing for unknown causes, disabling
differences and alternative routes to consciousness. Applied to
biological systems, it explains why analogical support generally weakens
with increasing causal distance from humans. Applied to contemporary AI,
it suggests that behavioural similarity provides only limited evidence
for consciousness because relevant causal correspondences remain poorly
established. The framework also clarifies what evidence would strengthen
claims of artificial consciousness.
\end{adjustwidth}
\vspace{14pt}
\end{@twocolumnfalse}
]

\section{Introduction}
Recent dramatic advances in artificial intelligence (AI), especially
Large Language Models (LLMs) and Large Multimodal Models (LMMs), have
sparked speculation that current or near-future AI models may be
conscious\supercite{butlin2023consciousness,butlin2026commentary,gurnee2026verbalizable,dehaene2026commentary}. Here we examine the question of whether
or not current work on AI provides a plausible route to the creation of
subjective experience -- ``what it is like to be
something''\supercite{nagel1974bat}. More specifically, our focus in this
paper is on \emph{subjectivity} -- what makes an entity feel some way at
all (as opposed to \emph{phenomenal character}, which refers to the
specific way it feels to be that entity\supercite{lee2026light}; or
\emph{access} consciousness, which refers to the availability of
information for reasoning, language, and control of
action\supercite{block1995confusion}).

In addition to its theoretical interest, the question of AI
consciousness has profound ethical and legal implications. Moral
philosophers, both religious and secular, generally acknowledge that any
being that is conscious (particularly if it experiences pain and
pleasure) has some moral value\supercite{singer1975animal,lee2025matter}. It has been argued
that sentient animals have certain inherent rights--perhaps they should
not be eaten by people or at least should not be mistreated and made to
suffer needlessly. If AI systems are conscious, then they too might
arguably have moral rights, perhaps even a claim to ``personhood''. If
accepted, such a conclusion would open up a vast array of thorny
questions (e.g., whether AI systems should be granted rights to own
intellectual or real property, or participate in the political system
along with humans\supercite{gunkel2017machine,solum1992legal,hull2023unlearning}).

\begin{figure*}[!t]
\centering
\includegraphics[width=\textwidth]{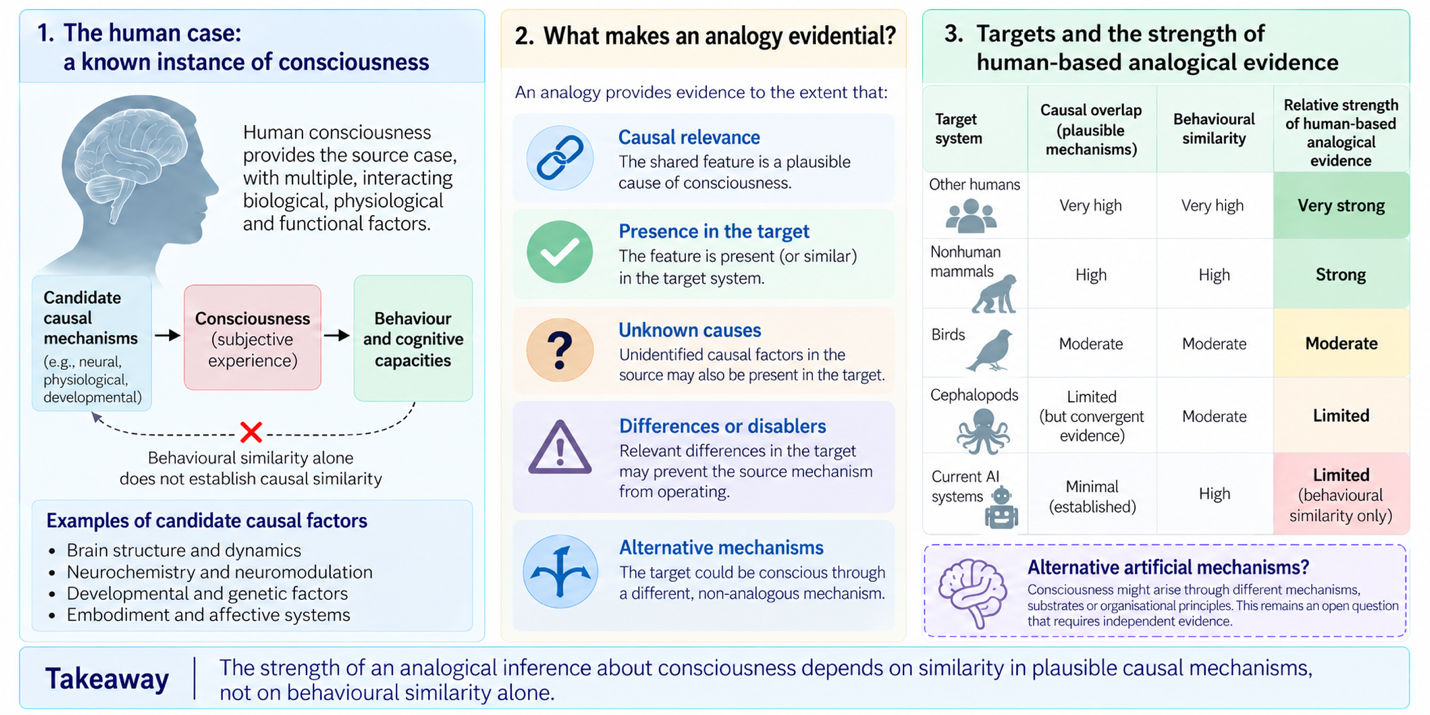}
\caption{A causal approach to evidential analogy for consciousness. Human consciousness provides the principal source analogue for evaluating consciousness in other systems. The framework distinguishes candidate causal features of consciousness from observable behavioural and cognitive outcomes, because similarity in downstream effects does not by itself establish similarity in the mechanisms that generate consciousness. The strength of an evidential analogy depends on the extent to which a target shares factors that are plausibly causally relevant, while also considering unknown causes, potential disablers and the possibility of alternative mechanisms. The right panel provides a qualitative, schematic comparison of different target systems; the categories are illustrative and not quantitative estimates.}
\label{fig:1}
\end{figure*}

\section{Evidential Analogy}

Because criteria for assessing consciousness in nonhuman animals have
implications for AI consciousness, we will also consider evidence
concerning the distribution of consciousness across animal species. Our
approach, summarized in Figure 1, is to examine the debate over AI
consciousness within a framework based on \emph{evidential analogy}. Our
aim is not to definitively
resolve
the debate (although we present our own assessment based on current
evidence), but rather to provide a theoretical framework to clarify the
empirical and logical issues at stake.

Beginning with Darwin, evolutionary biologists have used \emph{argument
by} \emph{analogy} in making the case for attribution of consciousness
to a range of vertebrates (for recent examples of such analogical
arguments, see Refs. \supercite{negro2026fetal,negro2025extrapolating}). In its most general form, an
argument by analogy makes an inferential claim: given that some known
\emph{source} is believed to have a certain characteristic or quality
\emph{Q}, and a novel \emph{target} is sufficiently similar to that
source in relevant ways, then the hypothesis that the target also has
\emph{Q} acquires evidential support. Here we examine the status of
arguments by analogy for the specific case in which \emph{Q} is
consciousness.

Analogy depends on similarity, but evaluating an analogical argument
requires defining \emph{which} similarities between source and target
are relevant to a given hypothesis about the target.

In the empirical realm of science, a general approach is to focus on
\emph{causal} \emph{relations}\supercite{hesse1963models,holyoak1985pragmatics,holyoak2025humanedge}: similarities
between known causes of effects observed in the source to potential
causes believed to operate in the target. A cause is a property that has
the power to bring about (or prevent) an effect\supercite{cheng1997covariation,hawthorne2001causal,pearl2011structural};
hence observing similar causes rationally supports an expectation of
similar effects. Although causes and their effects tend to cooccur, the
cause-effect relation is asymmetrical: effects do not have the power to
produce their causes\supercite{waldmann1992predictive}. If there is substantial
similarity of causal relations between source and target, an analogy
counts as \emph{evidential}: the source provides some degree of evidence
that the same or a similar effect occurs in the target.

Many analogies discussed in the context of science are \emph{not}
evidential. For the purpose of explanation or teaching, it is often
useful to point out relational parallels between a familiar source and a
less familiar target. For example, Rutherford's analogy between a solar
system and the structure of the atom (a standard example discussed by
cognitive scientists) is not evidential\supercite{holyoak1995mentalleaps}. The
underlying force that keeps planets revolving around the sun (gravity)
is not the same cause that keeps electrons revolving around the nucleus
(electromagnetic force). The analogy works well as a way of introducing
the theory of atomic structure to novices, but the fact that the sun
causes orbital motion of planets provides no evidence regarding how
atoms behave, because the causes are unlike. Rutherford's theory
garnered support based on experimental evidence, not from the analogical
comparison. In contrast, Newton's analogy between terrestrial and
celestial motion has evidential value because the same force (gravity)
causes both sets of phenomena.

As Rutherford's analogy illustrates, for pedagogical purposes it is
often best to find a source from a very different domain than the target
(i.e., a \emph{far} analogy), since the key requirement is that the
source be better understood than the target, and approximately
isomorphic to it (i.e., elements can be mapped one-to-one). But for an
analogy to be evidential, isomorphism is not sufficient. Rather, the
causal mechanism in the source must actually be the \emph{same} (at some
level of generality) as the corresponding target mechanism. It follows
that when the goal is to find an evidential analogy, \emph{near}
analogies are almost always the most powerful. This is particularly the
case when the set of relevant causal relations is unknown or poorly
understood, even in the source (as is currently the case for
consciousness). If the source and target are close to identical, we can
be confident that their relevant causal relations (whatever these may
be) are very likely shared. For example, because I know I have a mind
(endowed with inner experience), then by analogy I infer that other
humans also have minds, because they are like me in so many
ways\supercite{russell1948human}.

As summarized in Figure 1, similarities based on entities and relations
that are causally relevant should receive greater weight than
similarities based on downstream effects. Shared causal factors,
weighted by their importance, increase the strength of an analogical
argument; whereas causal differences (particularly those that could
disable a shared causal mechanism) decrease argument strength.

It is important to acknowledge the limitations of arguments by analogy.
Evidential analogies never guarantee that a hypothesis is correct.
Moreover, the \emph{absence} of a compelling evidential analogy does not
show conclusively that a hypothesis is false. When analogical support is
lacking, some other form of positive empirical evidence is required to
justify the hypothesis. However, we will show that under well-defined
conditions, the failure of a proposed analogy rationally decreases
confidence in a hypothesis that at least partly depended on it.

Despite their limitations, evidential analogies offer important
advantages. Most obviously, by carefully assessing causal structure, a
single source analogue can provide significant evidence in favor of a
hypothesis. Moreover, even when the causes of an effect are poorly
understood, high similarity involving \emph{possible} causes in the
source increases confidence that the target will exhibit effects similar
to those that occur in the source. In general, analogical reasoning is
particularly useful when examples are few and general laws have not been
established (e.g., in forming interpretations of archeological or
ethnographic observations\supercite{donnan1971ancient}). These conditions are
apparent when we examine the possible distribution of consciousness.

Box 1 presents a general method for applying evidential analogy to
debates about the distribution of consciousness. Our focus is on the
hypothesis that some target entity is conscious because it is similar to
humans (the source analogue known to be conscious). But our framework is
a general one, which can be applied to any scientific hypothesis based
in part or entirely on an evidential analogy.

To evaluate an evidential analogy, we need to ask not simply whether
source and target are similar, but whether they are similar with respect
to factors that plausibly contribute to the effect of interest. In the
present case, the source is human consciousness and the target may be
another animal species or an artificial system. Because the causal basis
of consciousness is almost certainly multi-factorial and remains only
partly understood, any formal treatment must accommodate both known and
unknown causal contributors.

Mackie\supercite{mackie1974cement} argued that real-world scientific phenomena
rarely stem from single necessary and sufficient causes. Rather, many
phenomena are multi-causal, depending on conjunctions of causes that
collectively yield an effect. Mackie introduced what he termed the
\emph{INUS condition}, which refers to a set of causes each element of
which is an \emph{I}nsufficient but \emph{N}ecessary part of an
\emph{U}nnecessary but \emph{S}ufficient set. An INUS set functions as
the actual and sufficient cause of an effect, in which each causal
element is necessary; but it allows for the possibility that there exist
other INUS sets, with different constituent elements, that could also
cause the same effect (i.e., $X_T$ in Eq. 1; see Box
1). Causal analyses based on variants of INUS sets have proved useful in
a broad range of applications, ranging from studies of social
innovation\supercite{ding2022innovation} to drug
administration\supercite{gwayichore2022defining} and medicine\supercite{thiem2022cora}.

In a strict INUS set, each constituent cause is a necessary element. The
overall probability of the effect is therefore a multiplicative function
of the probabilities of the individual elements, such that if any single
element is missing, the probability of the effect goes to zero. This
strict interpretation cannot be applied given our present state of
limited and uncertain knowledge about the causes of consciousness.
Indeed, the notorious Hard Problem of subjective
consciousness\supercite{chalmers1995facing} reflects the fact that some critical
causal elements remain unknown to science. In addition to this basic
epistemic uncertainty, strict INUS sets are unreasonable due to several
core biological features. These include neural plasticity (which enables
substitution of neural elements), nonlinear dynamics (consciousness may
be graded, or reflect a phase transition), and levels of abstraction
(reentrant signaling need not depend on any specific neuron).

We therefore treat the mechanism underlying human consciousness as a
\emph{soft INUS set}: a collection of candidate causal factors that may
differ in their importance and need not each be strictly necessary. This
formulation also allows for two additional possibilities. First, a
target may differ from the source in ways that disrupt an otherwise
similar causal mechanism. Second, a target might possess consciousness
through a mechanism that is not analogous to the human one at all. Box 1
summarizes the minimal formalism and translates it into five empirical
questions for evaluating a proposed analogy.

\begin{paperbox}{Box 1 \textbar{} A minimal framework for evaluating evidential analogies}
\small

Suppose we want to estimate whether a target system $T$ has some property $Q$,
such as consciousness, given that a source system $S$ is known to have $Q$. The
target could possess $Q$ either through a \textbf{source-like mechanism} $M_T$
or through some \textbf{alternative mechanism} $X_T$:
\begin{equation}
P(Q_T) \;=\; P(M_T)\,\bigl[\,1 - P(D_T \mid M_T)\,\bigr] \;+\; P(X_T).
\end{equation}
Here, $P(D_T \mid M_T)$ is the probability that some target-specific difference
disrupts an otherwise source-like mechanism.

Because consciousness is unlikely to depend on a single cause, we represent the
source mechanism as a \textbf{soft INUS set}: a collection of candidate causal
factors that may differ in importance and need not each be strictly necessary.
For each candidate factor $c$:
\begin{align*}
w_c &= \text{causal relevance of factor } c;\\
s_c(T) &= \text{similarity of target } T \text{ to the source on factor } c.
\end{align*}
The probability that the target instantiates the source-like mechanism can then
be represented as:
\begin{equation}
P(M_T) \;=\; P(U_T)\prod_{c}\bigl[\,1 - w_c\bigl(1 - s_c(T)\bigr)\,\bigr]
\end{equation}
where $P(U_T)$ represents uncertainty about whether unknown causal factors
present in the source are also present in the target.

Combining these components gives:
\begin{equation}
\begin{split}
P(Q_T) \;=\;
&\Bigl[\,P(U_T)\prod_{c}\bigl(1 - w_c(1 - s_c(T))\bigr)\Bigr]\\[2pt]
&\times\bigl[\,1 - P(D_T \mid M_T)\,\bigr] \;+\; P(X_T).
\end{split}
\end{equation}

The purpose of the formalism is not to assign precise numerical probabilities.
Rather, it identifies the kinds of evidence that should strengthen or weaken an
analogical inference. Similarity matters most when it concerns factors that are
independently plausible causes of the property of interest; similarity in
downstream behavioural or cognitive effects carries less evidential weight
unless those features are themselves causally relevant. The term $P(X_T)$ leaves
open the possibility that the target could possess $Q$ through a non-analogous
mechanism.

\begin{subbox}[Five questions for evaluating claims of consciousness]
\small
\qitem{1. Causal relevance}{\emph{Which features of the source are plausibly
involved in generating consciousness?} These factors should receive greater
causal weight $w_c$ than features that are merely correlated with, or
consequences of, consciousness.}

\qitem{2. Presence in the target}{\emph{To what extent are those causally
relevant features present in the target?} This is captured by $s_c(T)$, the
similarity between target and source on each candidate causal factor.}

\qitem{3. Unknown causes}{\emph{How much uncertainty remains about causal
factors that have not yet been identified?} $P(U_T)$ reflects the possibility
that unknown contributors to consciousness in the source are also present in the
target.}

\qitem{4. Relevant differences or disablers}{\emph{Does the target contain
differences that could interfere with an otherwise similar mechanism?}
$P(D_T \mid M_T)$ captures target-specific factors that could prevent the
source-like mechanism from producing consciousness.}

\qitem{5. Alternative mechanisms}{\emph{Is there independent evidence that
consciousness could arise through a different route?} Even if the target lacks a
human-like mechanism, $P(X_T)$ allows for consciousness generated by a
non-analogous mechanism, substrate or organizational principle.}
\end{subbox}

\begin{takeawaybox}
\small\noindent\textbf{Takeaway:} An evidential analogy is strong not when
source and target are simply similar overall, but when they are similar with
respect to plausible causes of the property being inferred. For consciousness,
behavioural resemblance can motivate the hypothesis, but causal resemblance
determines the strength of the human-based analogy.
\end{takeawaybox}

\end{paperbox}

The framework is intended primarily as a way of organizing evidence
rather than assigning precise numerical probabilities. Several
qualitative implications follow. First, \emph{near analogies are
especially informative when the true causal mechanism is incompletely
known.} If source and target are highly similar across many potentially
relevant dimensions, then even unknown causal factors are more likely to
be shared. As causal knowledge improves, however, overall resemblance
should matter less than similarity on those factors that are
independently shown to be causally important.

Second, \emph{known causal differences weaken an analogy in proportion
to their relevance.} A mismatch on an incidental feature should matter
little, whereas the absence of a factor with high causal weight should
matter much more. Likewise, target-specific features that interfere with
the operation of a shared mechanism can reduce analogical support.

Third, \emph{the framework does not assume that the human route to
consciousness is the only possible one}. Even if a target lacks a
human-like mechanism, consciousness could in principle arise through
some alternative pathway. Failure of the analogy therefore does not
establish absence of consciousness; rather, it shifts the evidential
burden to independent support for a non-analogous mechanism.

Finally, \emph{if a hypothesis initially gains support from a proposed
analogy, then evidence that the relevant causal overlap is weaker than
assumed should rationally reduce confidence in that hypothesis}. The
framework therefore provides a directional guide to inference even when
exact parameter values are unavailable.

\section{Consciousness in Humans}

Evaluating an evidential analogy requires careful consideration of the
causal factors operating in the source. Although the basis for human
consciousness is not entirely understood, decades of research have
coalesced into a relatively consistent empirical picture. Consciousness
seems to depend on the interactions between ascending arousal systems
originating in the brainstem, basal forebrain, hypothalamus, and the
thalamocortical system, with a key role of the so-called non-specific
thalamus\supercite{koch2016neural,aru2020cellular,storm2024integrative}, possibly well beyond the simple role
of gating cortical processing\supercite{coppola2025review}. Indeed, these are the
systems disabled by general anesthesia, across species, and which result
in network disruption of thalamocortical integration and frontoparietal
connectivity\supercite{lewis2012rapid,hudetz2016disconnecting,mashour2020conscious,monti2013dynamic}. This critical integration
regulates the sleep-wake cycle and generates the state transitions
(i.e., between wakefulness, NREM sleep, and REM
sleep\supercite{scammell2017neural}) and is known to be compromised in patients
with a Disorder of Consciousness\supercite{lutkenhoff2015thalamic}, in association
with the well-established electrophysiological signatures of depth of
impairment\supercite{lutkenhoff2020subcortical,lutkenhoff2022eeg}.

One aspect that is particularly important to highlight in this context
is that extensive empirical work shows that consciousness does
\emph{not} depend upon the presence of other aspects of cognitive
architecture. For example, consciousness can be preserved in the absence
of motor output in neurological conditions such as the locked-in
syndrome\supercite{laureys2005lockedin,kyselo2015lockedin,bruno2016consciousness}, during covert
awareness\supercite{owen2006detecting,monti2010willful}, and in MPTP-induced neurotoxic
lesions\supercite{langston1984pargyline}. Conversely, complex motor output can occur
without consciousness of the behavior, as in automatisms during complex
partial seizures\supercite{blumenfeld2012impaired}, somnambulism\supercite{castelnovo2018nrem},
and conditions such as alien hand syndrome\supercite{dellasala1991anarchic}.
Different forms of aphasia, including global, also do not appear to
interrupt the flow of experience (albeit changing its available
contents\supercite{nichelli2016aphasia,tononi2016neurology}), perhaps most strikingly in the case of
children with Angelman Syndrome, most of whom never develop any language
processing\supercite{andersen2001angelman}.

Furthermore, extensive language processing can occur without conscious
processing, including semantic activation from masked
words\supercite{dehaene1998imaging}, syntactic and multi-word processing under
continuous flash suppression\supercite{sklar2012reading}, and automatic
activation of translation equivalents in bilingual
individuals\supercite{thierry2007brain}. Finally, consciousness can be preserved
despite severe cognitive impairment. Patients with a frontal variant of
frontotemporal dementia lose executive function, social cognition, and
complex decision-making while remaining conscious\supercite{rascovsky2011sensitivity}.
Similarly, semantic dementia produces progressive loss of conceptual
knowledge without loss of consciousness\supercite{hodges2007semantic}, and severe
amnesic syndromes following medial temporal damage abolish episodic
memory formation while leaving consciousness intact\supercite{corkin2002hm}.

Conversely, it is well known that extensive complex cognitive processing
can occur in the absence of consciousness. Attention
processing\supercite{koch2007attention,dehaene2006conscious}, decision making\supercite{bechara1997deciding},
and even arithmetic and other complex cognitive
operations\supercite{sklar2012reading,kouider2007levels} can occur outside of consciousness
and/or operate on non-conscious items. In the context of the soft INUS
framework, such empirical evidence indicates that whether or not a
cognitive architecture is capable of the above cognitive functions
carries very little causal weight with respect to the probability of
possessing consciousness.

\section{Animals as a Calibration Case for Evidential Analogy}

Before we consider the prospects for artificial consciousness, it is
useful to review how the analogy with humans has guided inferences about
the distribution of consciousness across the only entities in the known
universe for which a rough scientific consensus has emerged: some
animals on planet Earth. Modern scientific explanation in biology is
founded on the theory of evolution, with the working assumption that all
characteristics of living things---including consciousness---are
products of evolution and its attendant selection pressures. Within the
biological case, the best-established candidate causes of consciousness
are neurobiological and physiological processes within the organism. In
the context of biology, evidential analogy is guided by analyses of
cause-effect relations. Broadly speaking, the structure of an organism
provides the inner ``machinery'' that causes its outward effects,
including its behavioural phenotype. For example, a specific pattern of
neural receptors or a specific clearance rate for a neurotransmitter
might enable problem-solving behaviors indicative of high intelligence.
With respect to evidential analogy, the key point is that causal weights
will be highest for factors involved in the biological mechanism that
produces consciousness, and lower for aspects of the biological
phenotype (such as complex problem solving) that themselves appear to be
effects of biological mechanisms, rather than causes of consciousness.

A scientific consensus on the distribution of biological consciousness
was provided by the Cambridge Declaration on
Consciousness\supercite{low2012cambridge}, which summarizes evidence that forms
of inner experience are present in (at least) all mammals and birds, as
well as some cephalopods, notably the octopus.\footnote{Others have
  argued that evidence supports broader distribution of consciousness
  across taxa (perhaps encompassing all vertebrates as well as some
  cephalopods and some arthropods, including insects and
  crustaceans\supercite{ginsburg2019evolution}).} In broad strokes, this scientific
consensus attributes consciousness to species that share a general
biological phenotype, which includes a sleep/wake cycle coupled with
evidence of dreaming during sleep\supercite{penaguzman2023animals}, as well as
similar drug sensitivities. Nonhuman species that meet these general
criteria for consciousness are among the more intelligent
animals, though none exhibit human-level reasoning or linguistic
abilities. Based on evolutionary history, there is clearly a gradient of
overall similarity that runs from humans to other primates, non-primate
mammals, birds, and cephalopods. Homology to humans is essentially
complete for great apes (common ancestor \textasciitilde6.4 million
years ago, MYA; all divergence estimates drawn from
TimeTree\supercite{kumar2022timetree}). The nervous system organization, sensory
apparata, body plan, and behavioral repertoire of primates resemble
those of humans at nearly every level of description.

Homology is also extensive for other mammals (common ancestor with mice
\textasciitilde87 MYA), for which brain organization, sensory systems,
and neurochemistry are broadly conserved even as body plans and
behavioral repertoires diverge. The strength of the homology is based on
genetic pathways, shared developmental machinery, and shared
neurochemical systems. For example, mice and humans respond to the
anesthetic propofol with similar behavioral and physiological signatures
of unconsciousness (a sleep-like state), including immobility,
analgesia, and amnesia\supercite{brown2010general,mashour2022anesthesia}. A similar case can be
made for sleep: mice and humans share sleep architecture, characteristic
electrophysiological signatures, autonomic and hormonal patterns, as
well as wake-promoting and sleep-promoting neural
circuits\supercite{scammell2017neural}. This multi-dimensional convergence,
underwritten by homology, is what makes the case for consciousness in
mammals particularly strong.

Homology is partial for birds (common ancestor with mammals
\textasciitilde319 MYA, in the amniote lineage). Though birds share deep
brainstem and mid-brain homology with mammals, their forebrain
neuroanatomy is radically different (nuclei versus a laminar, layered
organization), although in some respects functionally
similar\supercite{gunturkun2016cognition,prior2008mirror}. Finally, homology is minimal for
cephalopods (last common ancestor with vertebrates \textasciitilde686
MYA). Cephalopods evolved large, distributed nervous systems
independently of the vertebrate lineage. Their invertebrate body plan
and independently-derived sensory apparata diverge from the vertebrate
template at most levels of description\supercite{godfreysmith2013cephalopods,mather2019octopus,shigeno2018cephalopod}. Despite
these important differences, mammals, birds, and cephalopods exhibit
functional correspondences between brain regions that appear to support
consciousness.

The animal cases demonstrate that analogical support need not depend on
superficial resemblance or human-level intelligence. Rather, evidential
analogy depends on convergent evidence that the target shares structures
and processes plausibly related to consciousness.

\section{What Does the Human Analogy Imply for AI Consciousness?}

What evidential support does the human analogy provide for consciousness
in current advanced AI systems or their near-future successors? We refer
here primarily to LLMs and LMMs implemented as artificial neural
networks, most commonly based on a transformer
architecture\supercite{vaswani2017attention}. However, our analysis can readily be
extended to all AI systems and computer simulations of cognition that
fall loosely in the tradition of abstract Turing
machines\supercite{holyoak2024turing}, including those that are explicitly
symbolic or rule-based, as well as hybrid neurosymbolic systems (for a
recent overview, see Ref. \supercite{colelough2025neurosymbolic}).

As always when consciousness is considered, the key evidential source
analog is the case of humans. Indeed, this is particularly true for AI
systems, as their behavioral similarities (primarily linguistic) are
entirely with humans (and not, for example, with other species likely to
be conscious, such as the octopus). However, these behavioural
similarities have not been established as causes of consciousness and
are more naturally treated as downstream manifestations of the
mechanisms that generate cognition. Most current AI systems are based on
artificial neural networks, but their resemblance to biological neurons
is limited to the fact that both can be viewed as implementations of
logic gates. None of the internal complexities of biological neurons
(see Ref. \supercite{cao2022multiple}) carry over to their artificial counterparts.
Given the uncertainty about the true INUS set of causes that yield human
consciousness, non-zero causal weights must be assigned to an immense
set of potential neurobiological factors---genetic, epigenetic,
developmental, neurochemical, electromagnetic, architectural---that AI
systems lack altogether. At present there is no independent evidence
that the unknown factors involved in human subjectivity are instantiated
in these systems.

Given that the function for integrating causal probabilities in an INUS
set is multiplicative, a sufficiently large mismatch on a high-weight
factor can sharply reduce the probability of a shared mechanism. The
qualitative conclusion is that the many causally-relevant
dissimilarities between humans and AI systems make the expected value of
\emph{P}(\emph{M\textsubscript{T}}) low. Physical substrate differences
therefore remain an unresolved source of uncertainty about whether a
human-like consciousness-generating mechanism could be realized in
present hardware\supercite{cao2022multiple,seth2025biological}.

Many proponents of potential AI consciousness ground their case on the
well-known assumption of \emph{computational functionalism} that has
dominated both AI and cognitive science for decades. This philosophical
position\supercite{putnam1960minds,fodor1975language} asserts that mental states are defined
by their computational roles rather than their physical makeup. Often
summarized by the slogan "the mind is the software of the brain," this
framework implies that any physical system, whether constructed from
biological tissue or silicon chips, can possess a mind (including
conscious experience) if it runs the right program. If functionalism
holds, one could claim that the host of neurobiological factors that
distinguish human brains from AI systems can be confidently excluded
from the true INUS set for human consciousness. The only factors
relevant to consciousness, either in humans or computers, would be the
necessary abstract computer program.

However, implementing computational features associated with theories of
conscious access has not, by itself, supplied evidence for subjective
experience. Notably, several computer implementations of the global
workspace theory have been developed\supercite{baars2007architectural,dehaene2011global,franklin2009evolutionary,madl2011timing}. These
simulations were intended (and interpreted by their authors) as tests of
a theory of access consciousness, providing neural-style implementations
of attention, working memory, and global broadcasting. The computer
programs achieved substantial success in capturing a broad array of
neural and behavioural findings from research on humans. However, the
researchers explicitly did \emph{not} conclude that their computer
simulations of access consciousness had acquired inner experience, and
no evidence of such an outcome was reported.

More recently, there has been cautious speculation that for some LLMs,
their extensive training has produced internal representations that have
some of the properties associated with a global
workspace\supercite{gurnee2026verbalizable}. Very roughly, analytic techniques have
been used to identify specific vectors within the model that have the
potential to be ``verbalizable'' in the sense of being similar to the
vectors for tokens (words or parts of words) in the model's vocabulary.
Whether or not these findings satisfy the requirements of global
workplace theory is debatable\supercite{butlin2026commentary,dehaene2026commentary}. But in any case,
identifying vectors in an LLM that approximate words does not constitute
evidence that the LLM has any sort of inner experience. Since the dawn
of AI, symbolic models have been explicitly constructed using
``verbalizable'' internal representations (e.g., an internal symbol for
``cat''), but no one has seriously argued such models are conscious.
Recent work on mechanistic interpretability in LLMs is of considerable
interest in demonstrating that training these models may yield
symbol-like internal representations, but this possibility provides no
evidence that the models have acquired inner experience.

The similarities that justify serious consideration of the hypothesis
that AI could be conscious are almost entirely based on the fact that AI
systems have recently shown evidence of behaviours that equal or exceed
human performance levels across a broad range of intellectual tasks. But
the key question is not whether there are many similarities between
humans and AI; rather, the question is whether any causal factor
invalidates the causal mechanism in the source as a potential mechanism
in the target. For biological systems, phenotypical behavior is best
viewed as the effect of underlying neurobiological mechanisms.
Intelligent reasoning is plausibly an \emph{effect} of conscious
experience, or of the mechanism that produces it, but it is not a
plausible \emph{cause} of consciousness. Hence behaviors such as writing
stories or coding websites should be assigned low causal weights.

Notably, other species that appear to be conscious, such as birds and
octopi, lack human-level intellectual abilities, though they share more
universal aspects of the biological phenotype for consciousness
(sleep/wake cycle, drug sensitivities). In the biological case,
consciousness appears to have evolved hundreds of millions of years
prior to high-level cognitive abilities, some of which appear to be
unique to the (evolutionarily young) human species\supercite{penn2008darwin}.
Faith that building smarter and larger AI systems will inevitably yield
inner experience can be considered a variety of the ``intellectualist
fallacy'' first criticized by James\supercite{james1909meaning}. In the present
context, the fallacy is that ``intelligence'', broadly construed, is the
only or most important aspect of what it means to be human; and that in
particular, high intelligence is sufficient to generate conscious
experience. In biological evolution, consciousness appears to have
preceded the emergence of the high-level reasoning and linguistic
capacities characteristic of humans.

Finally, the evidential analogy from human to AI consciousness is
undermined by the fact that we know the basic cause of AI behaviour very
well. The entire point of AI is to engineer human-like intellectual
abilities. Current AI systems are trained extensively on large corpora
of human-generated linguistic and multimodal data, with the explicit
goal of producing useful, human-interpretable behaviour. To a very
impressive extent, AI researchers have succeeded in engineering
human-like behaviour using an entirely non-human method. There is
currently little evidence that the biological mechanisms implicated in
human consciousness play a corresponding causal role in the mechanisms
by which AI systems generate human-like behaviour.

As we acknowledged at the outset, our analysis of the evidential analogy
from human to machine consciousness is by no means definitive. New
scientific knowledge about the biological mechanism for consciousness
might resurrect the analogy. But if the human-like pathway contributes
little analogical support, then the case for AI consciousness
increasingly depends on $P(X_T)$: evidence
for an alternative route. Estimates of the probability that
non-biological consciousness is possible hinge on the prior that
alternative routes to consciousness exist. These priors will vary
widely. One person might argue that the only known form of consciousness
in the entire universe is the biological kind found in various species
of animals on our lonely planet, so $P(X_T)$
is extremely low. Someone else might call attention to the fact that the
universe is large indeed, admonish us to eschew ``biological
chauvinism'' or ``carbon chauvinism'', and claim that
$P(X_T)$ is extremely high. Whatever its
outcome, this debate is entirely different from that about whether AI
might be conscious because it does things humans can do, only better.

Our evaluation of the potential for AI consciousness has been limited to
silicon-based systems in the tradition of abstract Turing machines. Even
if our negative conclusion about their potential for consciousness
stands, and even if consciousness is in important ways dependent on a
biological substrate, it remains possible that it could be artificially
created. As noted by Searle\supercite{searle2000consciousness}, the biological basis
for consciousness might someday be replicated, perhaps in androids
constructed from organoids. Such systems would differ importantly from
conventional silicon-based AI because they might reproduce components of
the biological pathway itself, rather than relying on an entirely
alternative mechanism. The moral implications of this possibility raise
profound ethical questions.

Box 2 summarizes the empirical research agenda implied by this
framework. Progress will require better identification of the causal
basis of consciousness in known biological systems, more principled
comparative inference across species, and explicit tests of both
human-like and alternative mechanisms in artificial systems.

\begin{paperbox}{Box 2 \textbar{} Key research questions for evaluating consciousness beyond humans}
\small

\begin{subbox}[1. Clarifying the human source case]
\small
Progress in comparative inference depends first on identifying which features of
human biology are actually involved in generating subjective experience.
\begin{itemize}[leftmargin=1.1em,itemsep=2pt,topsep=4pt]
\item Which biological and computational processes are causally necessary,
      enabling, or merely correlated with consciousness?
\item Which candidate mechanisms survive intervention-based tests, including
      lesion, anaesthesia, and perturbation studies?
\item How should uncertainty about still-unknown causes of human consciousness
      be represented when evaluating other systems?
\end{itemize}
\end{subbox}

\begin{subbox}[2. Calibrating analogies across animals]
\small
Comparative biology provides the best-developed test case for extending
consciousness beyond humans, but similarities must be weighted by their causal
relevance.
\begin{itemize}[leftmargin=1.1em,itemsep=2pt,topsep=4pt]
\item Which neural, physiological and behavioural similarities to humans provide
      the strongest evidence for consciousness across species?
\item How should shared ancestry and convergent evolution be weighed when
      assessing causal correspondence?
\item Which biological differences meaningfully weaken an analogy, and which are
      likely to be incidental?
\item Can comparative evidence identify features that are sufficient for
      consciousness even when human-like neuroanatomy is absent?
\end{itemize}
\end{subbox}

\begin{subbox}[3. What would count as evidence for AI consciousness?]
\small
Behavioural competence alone cannot determine whether an artificial system is
conscious. Stronger evidence would need to connect properties of the system to
mechanisms plausibly involved in generating subjective experience.
\begin{itemize}[leftmargin=1.1em,itemsep=2pt,topsep=4pt]
\item Which computational or architectural properties, if any, can be shown to
      play a causal role in consciousness rather than merely supporting
      intelligent behaviour?
\item How can mechanistic evidence for consciousness be distinguished from
      behavioural imitation or successful prediction of human reports?
\item What perturbational or intervention-based tests could reveal whether
      candidate AI mechanisms play roles analogous to consciousness-related
      mechanisms in biological systems?
\item What evidence would justify increasing confidence in consciousness even
      when an AI system differs substantially from human biology?
\end{itemize}
\end{subbox}

\begin{subbox}[4. Alternative routes and emerging systems]
\small
Weak evidence for a human-like mechanism does not exclude consciousness arising
through some different causal route. Such claims, however, require independent
evidence.
\begin{itemize}[leftmargin=1.1em,itemsep=2pt,topsep=4pt]
\item What empirical findings would support a genuinely non-biological route to
      consciousness?
\item How can substrate-neutral theories of consciousness generate predictions
      that distinguish conscious from non-conscious systems?
\item Which properties of neuromorphic, organoid and hybrid
      biological--artificial systems are most relevant to evaluating
      consciousness?
\item At what point would an emerging system cease to be a distant analogue of
      the human case and become a new source case for future comparisons?
\end{itemize}
\end{subbox}

\begin{takeawaybox}
\small\noindent\textbf{Takeaway:} Progress will require moving from behavioural
resemblance to causal tests: identifying mechanisms that generate consciousness
in known biological systems, determining how those mechanisms generalize across
species, and specifying what independent evidence would support genuinely
alternative routes to conscious experience.
\end{takeawaybox}

\end{paperbox}

\section{Conclusion}

Questions about the distribution of consciousness cannot be resolved by
direct observation alone. Because subjective experience is directly
accessible only in the first person, inferences about consciousness in
other entities must rely in part on analogy. Our central claim is that
not all analogies are equally informative: the evidential force of an
analogy depends primarily on similarity in factors that are plausibly
involved in generating consciousness, rather than on similarity in
downstream behavioural or cognitive effects.

This distinction helps explain why the human analogy supports
consciousness in many animals more strongly than it supports
consciousness in current AI systems. Across biological species,
evolutionary continuity, shared physiology and convergent functional
organization provide at least some basis for causal comparison with the
human case. By contrast, although contemporary AI systems exhibit
impressive behavioural similarity to humans, their correspondence to the
candidate mechanisms currently thought to underlie human consciousness
is much weaker. For that reason, the human-based analogy provides only
limited support for attributing consciousness to present AI.

At the same time, weak analogical support from the human case should not
be confused with proof of impossibility. Artificial consciousness could
in principle arise through a different mechanism, substrate, or
organizational route. The key scientific challenge, therefore, is not
simply to ask whether AI behaves like us, but to identify what kinds of
evidence would justify claims about consciousness in any non-human
system. Progress will depend on clarifying the causal basis of
consciousness in humans, improving comparative evaluation across animal
taxa, and developing principled criteria for assessing genuinely
alternative routes to conscious experience.

\section{References}
{\small
\printbibliography[heading=none]
}

\end{document}